\pdfoutput=1

\documentclass[11pt,a4paper]{article}

\usepackage[hyperref]{ranlp2021}
\usepackage{times}
\usepackage{latexsym}

\usepackage{times}
\usepackage{latexsym}
\usepackage{afterpage}
\aclfinalcopy

\usepackage{adjustbox}

\newcommand\blfootnote[1]{%
  \begingroup
  \renewcommand\thefootnote{}\footnote{#1}%
  \addtocounter{footnote}{-1}%
  \endgroup
}

\usepackage[T1]{fontenc}

\usepackage[utf8]{inputenc}

\usepackage{microtype}

\usepackage[disable]{todonotes}
\usepackage{booktabs}       
\usepackage{amsfonts}       

\title{Czert -- Czech BERT-like Model for Language Representation}

\def \itemizeseparatorsize {0em}

\author{Jakub Sido\textsuperscript{*} \and   Ondřej Pražák\textsuperscript{*} \and {Pavel Přibáň}\\ \and \bf{Jan Pašek} \and Michal Seják \and Miloslav Konopík\\[0.5em]
\tt{\{sidoj,ondfa,pribanp,pasekj,sejakm,konopik\}@kiv.zcu.cz}\\[0.5em]
 NTIS -- New Technologies for the Information Society, \\
Department of Computer Science and Engineering, \\
Faculty of Applied Sciences, University of West Bohemia, Technick\'a 8, 306 14 Plze\v{n} \\
Czech Republic
}

\makeatletter
\newcommand\footnoteref[1]{\protected@xdef\@thefnmark{\ref{#1}}\@footnotemark}
\makeatother

\begin{document}
\maketitle
\begin{abstract}
This paper describes the training process of the first Czech monolingual language representation models based on BERT and ALBERT architectures. We pre-train our models on more than 340K of sentences, which is 50 times more than multilingual models that include Czech data. We outperform the multilingual models on 9 out of 11 datasets.  In addition, we establish the new state-of-the-art results on nine datasets. At the end, we discuss properties of monolingual and multilingual models based upon our results. We publish all the pre-trained and fine-tuned models freely for the research community. 
\end{abstract}

\section{Introduction}
\blfootnote{
    %
    %
    %
    %
    \hspace{-0.65cm}\textsuperscript{*}Equal contribution.}
Transfer learning and pre-trained word embeddings became a crucial component for most Natural Language Processing (NLP) models. Contextualized methods \cite{cove,peters-etal-2018-ELMO,howard-ruder-2018-ULMFIT} overcame the initial context insensitive word embeddings approaches \cite{mikolov2013distributedword2vec,pennington-etal-2014-glove,bojanowski2017enriching}.  \cite{cove,peters-etal-2018-ELMO}. The word representations generated by the named methods are usually used as input features for other task-specific models that are further trained. Starting with the BERT \cite{bertpaper}, the BERT-like models \cite{lan2019albert,liu1907roberta,sanh2019distilbert,nips-2019-XLNET} based on Transformer architecture \cite{attention-all-transformer}, achieved a significant performance improvement in many NLP tasks \cite{raffel2019exploring}. These recent models are trained on a language model task or tasks that are closely related to it. Such pre-training allows them to capture the general representation of language and text. The pre-trained models are then directly fine-tuned with specific data for a selected downstream task. The performance improvement of these models is paid by the vastly increased requirements (i.e., data and computational resources) for their training.

\par The mentioned models are primarily trained for English.
Recently, models for other, mostly larger, languages have been released, e.g., French \cite{martin2019camembert,le2019flaubert}, Polish \cite{Kleczek2020-polbert}, Turkish \cite{stefan-schweter-BERTurk}, Russian\footnote{\url{http://docs.deeppavlov.ai/en/master/features/models/bert.html}}, Italian\footnote{\label{note1}\url{https://github.com/dbmdz/berts}}, German\footnoteref{note1}, Arabic \cite{safaya2020kuisail-arabicbert}, but also for languages that are spoken by a relatively small number of people, i.e.,  Romanian \cite{dumitrescu2020birth-romanianBERT}, Dutch \cite{dutch-BERT} or Finish \cite{finish-bert-2019}. There were also introduced multilingual models \cite{nips-2019-XLM,conneau-etal-2020-unsupervised-XLM-Roberta}, that can be used for multiple languages at once but usually at the cost of lower performance in comparison to solely monolingual models \cite{martin2019camembert,finish-bert-2019,dumitrescu2020birth-romanianBERT} as we show in this paper.


\par Our main motivation is to train and provide publicly available models\footnote{The model is available at \url{https://github.com/kiv-air/Czert}} for the Czech language that performs better than available multilingual models.

\par In this paper, we describe a process of training of two BERT-like models for Czech language and their evaluation on six tasks along with a comparison to two multilingual models, i.e. mBERT \cite{bertpaper} and SlavicBERT \cite{arkhipov2019tuning-SlavicBert}. More concretely, the architectures of our models are based on the ALBERT \cite{lan2019albert} model (Czert-A) and the original BERT \cite{bertpaper} model (Czert-B). Both of our models are trained on a text corpus of the approximate size of 36 GB of plain text consisting of Czech Wikipedia articles, crawled Czech news and Czech National Corpus \cite{cs-nat-corpus}. We train the models from scratch (i.e., with random initialization) using \textit{Masked Language Model} (MLM) and \textit{Next Sentence Prediction} (NSP) tasks as training objectives with a slight modification of the NSP task, see Section \ref{sec:pretraining}. We evaluate our models on six tasks\footnote{Some of the evaluation tasks contain more than one independent dataset.}: Semantic Text Similarity (STS), Named Entity Recognition (NER), Morphological Tagging (MoT), Semantic Role Labeling (SRL), Sentiment Classification (SC) and Multi-label Document Classification (MLC).

\par Our main contributions are the following ones: 1) We release a pre-trained and ready to use BERT model (Czert-B) for the Czech language that outperforms the compared models on all evaluated sentence-level tasks and it performs comparably on Semantic Role Labeling task. Along with the pre-trained model, we also release the fine-tuned models for each task. 2) We achieve new state-of-the-art results on seven datasets. Moreover we outperform the multilingual models with our newly trained Czert-B model on 7 out of 10 datasets.





\section{Related Work}
\label{sec:related_work}
\subsection{English BERT and ALBERT}\label{sec:bert}

The BERT \cite{bertpaper} model adopts the multi-layer Transformer-encoder architecture \cite{attention-all-transformer} with two pre-training tasks: \emph{Masked Language Modeling} and \emph{Next Sentence Prediction}.

The goal of the \emph{MLM} task is to recover artificially distorted sentences where some of the original tokens are \emph{masked out} (hidden), and some are randomly \emph{replaced} with other tokens. These distorted tokens and few other unchanged tokens are selected for prediction (classification). The ratios of predicted tokens can be tuned. For example, in the original BERT model, 15\% of input tokens are predicted, 80\% of them are masked out, 10\% are changed randomly, and 10\% are left intact.

The \emph{NSP} is a binary classification task of sentence pairs. For two sentences \texttt{A} and \texttt{B} taken from the training corpus, the goal is to decide whether the sentence \texttt{B} is the actual next sentence (following the sentence \texttt{A}) or whether it is a randomly selected sentence from the corpus.
In the BERT paper \cite{bertpaper}, the random sentences are sampled uniformly from the whole corpus.

The BERT model represents a big step in massively pre-trained models. The experiments\footnote{Experiments in the BERT paper \cite{bertpaper} or in many consequent research papers.} show that a large stack of cross-attention layers with a huge amount of parameters of BERT and BERT-like models can significantly boost the performance of many downstream tasks. A relatively short fine-tuning phase is usually sufficient to set new state-of-the-art results in many tasks using the pre-trained model.


In the original paper \cite{bertpaper}, the authors publish the BERT\textsubscript{BASE} and BERT\textsubscript{LARGE} models. BERT\textsubscript{BASE} contains 12 layers, 12 attention heads, and the size of the hidden state is set to 768. In total, it requires 110M parameters. The BERT\textsubscript{LARGE} model has 24 layers, 16 attention heads and the size of the hidden state is set to 1024, which results in 340M parameters.

Training such huge models requires vast computational resources. Therefore, researchers developed methods to reduce the training complexity, memory demands or prediction time, while maintaining similar performance on the fine-tuned tasks. ALBERT model \cite{lan2019albert} represents an example of such an approach. 


ALBERT slightly modifies BERT to use the parameters more effectively. First, the authors argue that word embedding size equal to the hidden size (768 for base) is unnecessarily large. They propose to use a smaller size (128) and project the embeddings to the hidden size, which significantly reduces the number of parameters (25M less than in the base variant). Another modification is in cross-layer parameter sharing. In ALBERT, all the weights are shared across all the layers. Another modification consists of replacing the NSP task with a harder task of sentence ordering prediction (SOP). 
That should result in making the model understand semantics better. The authors introduce models ALBERT\textsubscript{BASE}, ALBERT\textsubscript{LARGE}, ALBERT\textsubscript{XLARGE}, ALBERT\textsubscript{XXLARGE} with 12M, 18M, 60M and 235M parameters, see Table \ref{tab:params}.


\subsection{ BERT-like Models for Other Languages}

Researchers publish a multilingual variant of standard BERT\textsubscript{BASE} model (\emph{mBERT})\footnote{See \url{https://github.com/google-research/bert/blob/master/multilingual.md}.}. It is jointly trained on Wikipedia pages of 104 languages. The model settings are almost the same as in BERT\textsubscript{BASE}; it differs only in the vocabulary size\footnote{BERT\textsubscript{BASE} uses a vocabulary with 30K sub-word tokens while mBERT increases the size to 120K tokens.}. 

However, researchers around the world trained the monolingual variant of the BERT and showed the domination of the monolingual version over the mBERT in many tasks, for example, French \cite{martin2019camembert}, Finish \cite{finish-bert-2019} or Romanian \cite{dumitrescu2020birth-romanianBERT}.  


\citet{arkhipov2019tuning-SlavicBert} used a combination of four Slavic languages: Bulgarian, Czech, Polish, and Russian. They trained their model using Wikipedia dumps for all four languages and a huge set of Russian news texts.
They use the same model architecture and training process as mBERT, and they initialized the model with mBERT weights.

\begin{table}[ht!]
    \centering
\begin{adjustbox}{width=\columnwidth,center}
    \begin{tabular}{lcccc}
    \toprule
         &  BERT\textsubscript{BASE} & ALBERT\textsubscript{BASE} & mBERT & Slavic BERT\\ 
         \midrule
        Params & $110M$ & $12M$ & $170M$ & $170M$ \\
        Vocab size & $40K$ & $40K$ & $120K$ & $120K$ \\
        Emb. params & $\approx30M$ & $\approx5M$ & $\approx90M$ & $\approx90M$ \\  
        \bottomrule
    \end{tabular}
    \end{adjustbox}
    \caption{Related models parameters.}
    \label{tab:params}
\end{table}

\section{Pre-training Process}
\label{sec:pretraining}
\subsection{Dataset Description}
Training BERT-like models require to collect large quantities of raw text data, pre-process them and prepare automatically labeled training data. 

\paragraph{Training corpora}
We use two publicly available corpora and our crawled dataset of Czech news:
\begin{itemize}
    \setlength\itemsep{\itemizeseparatorsize}
    \item Czech national corpus (\emph{CsNat}) \textit{28.2GB}, \cite{cs-nat-corpus},
    \item Czech Wikipedia (\emph{CsWiki}) \textit{0.9GB}, dump\footnote{Taken from \url{https://dumps.wikimedia.org}} from May 2020,
    \item Crawled of Czech news (\emph{CsNews}), \textit{7.8GB}.
\end{itemize}

The \emph{CsNat} corpus composes of randomly-ordered blocks of texts sized maximum size of 100 tokens. Each block contains at least one sentence.
This must be considered later for the NSP task, which requires a continuous block of texts. Table \ref{tab:corpora_sizes} shows the sizes of each corpus in terms of blocks and sentences counts.

\paragraph{Pre-processing}

We prepare two versions of the corpus: \emph{cased} and \emph{uncased}. Both versions are tokenized with the \emph{WordPiece} tokenizer \cite{wu2016google} which is trained on the entire corpus.

\paragraph{Pre-training Objective}
We employ MLM and NSP tasks (see section \ref{sec:bert}) for training our model.

The MLM task is used exactly as in the BERT model. The NSP task needs a few considerations. The NSP task requires the availability of continuous blocks of text to form pairs of sentences where one sentence follows the other. At the end of each block, we lose the last sentence that has no sentence to form a pair with. The effect of this issue becomes more apparent with the decreasing length of the continuous text blocks, such as in the case of the \emph{CsNat} corpus. Here, we observe 5.6 sentences per continuous block on average. That means that we are able to use 4.6 sentences out of 5.6 (i.e. approximately 18\% of sentences cannot form a pair). When compared to the two remaining corpora, this number is relatively high. In the \emph{CsWiki} and \emph{CsNews} corpora, only 6\% and 4\%, respectively, of sentences cannot form a pair.

Moreover, we design more difficult negative samples for the NSP task -- we select sentences from the same paragraph (that do not directly follow the first sentence) to build non-trivial negative pairs instead of drawing random sentences from the whole corpora as in BERT.

The final dataset consists of 578 158 196 training pairs of sentences. In Table \ref{tab:corpora_sizes}, we provide some basic statistics of the dataset used in our setup. 





\begin{table}[ht!]
\centering

\begin{adjustbox}{width=\linewidth,center}
\begin{tabular}{lccc}
\toprule

& Textual Blocks & Sentences & Avg/block  \\ \midrule
CsNat  & 49 104 507& 275 314 224 &   5,61 \\ 
CsWiki & 450 000   &6 964 794 &   15.48 \\ 
CsNews & 2 625 306 &  58 979 893 & 22.47 \\ 

\bottomrule
\end{tabular}
\end{adjustbox}

\caption{Statistics of coropra used.}
\end{table}\label{tab:corpora_sizes}

\subsection{Models}
We train two models: a smaller ALBERT\textsubscript{BASE} model (Czert-A, 12M parameters) and a larger BERT\textsubscript{BASE} model (Czert-B 110M parameters).

\paragraph{Czert-A}
is very similar to the standard ALBERT\textsubscript{BASE} with a few modifications:
we use WordPiece tokenizer, 
the batch size is set to 2048 (due to cluster limits), and we use our version of NSP introduced in Section \ref{sec:pretraining} instead of SOP.
\paragraph{Czert-B} is configured exactly as the BERT\textsubscript{BASE} model with increased batch size to 2048.

\paragraph{Optimization}
Both models are trained using a learning rate of 1e-4 with the linear decay using Adam optimizer \cite{Kingma-adam}. First, we iterate over the dataset once (single epoch) with the maximum sequence length set to 128. It leads to 300K batches (steps). Similarly to the BERT approach, we then increase the maximum sequence length to 512. We perform about 50K
steps with the increased sequence length. In this second shorter iteration, we decrease the batch to 256 samples to fit the cluster memory limits. More details about the computational cluster and its configuration are located in Appendix \ref{sec:cluster}.





\section{Evaluation}
The following section summarizes the performance of Czert on various tasks and compares our model with similar available models. We also add experiments without the pre-training phase to highlight the impact of additional unsupervised data in the Czech language. We also compare Czert with the following baselines:

\paragraph{Baselines}
\begin{itemize}
    \setlength\itemsep{\itemizeseparatorsize}    
    \item \textit{SlavicBERT} -- a model trained on four Slavic languages (Russian, Bulgarian, Czech and Polish)\cite{arkhipov2019tuning-SlavicBert},
    
    \item \textit{mBERT} -- a multilingual version of BERT \cite{bertpaper},
    
    \item \textit{ALBERT-r} -- a randomly initialized ALBERT model without any pre-training.
    
\end{itemize}

\subsection{Evaluation Tasks}\label{sec:tasks}

We evaluate our models on six tasks that cover three main groups of NLP tasks:
\emph{Sequence Classification} (Sentiment Classification, Multi-label Document Classification); \emph{Sequence Pair Classification} (Semantic Text Similarity); \emph{Token Classification} (Morphological Tagging, Named Entity Recognition, Semantic Role Labeling)



For the \emph{sequence classification} tasks, we take the \emph{pooled} output of the BERT model (and ALBERT). We add dropout and an output layer. The number of output neurons and the activation function differs for each task. 


\emph{Sentence pair classifications} tasks employ the same approach as sequence classification tasks. The only difference is that we feed both sentences separated with special \texttt{[SEP]} token together into the model. This way, the model can profit from \emph{cross-attention} between tokens from different sentences.

For the \emph{token classification} tasks, we use the output embeddings 
associated with the input words (\texttt{[CLS]}, \texttt{[SEP]} and other special output embeddings are ignored). When the input words are split to sub-word tokens, we take only the first sub-word tokens. 
For optimization, we use the \emph{Cross-entropy} loss.

For all the tasks, the newly added layers are initialized randomly. We employ the \emph{Adam} optimizer.

\subsection{Named Entity Recognition}\label{sec:T_NER}

We use two different datasets to evaluate our model on the named entity recognition task. These are the following:

\begin{enumerate}
    \item \textbf{Czech Named Entity Corpus} (CNEC) \cite{SevcikovaEtAl2007CNEC} containing 4 688 training, 577 development and 585 test sentences. We use the CoNLL version of the dataset \cite{CRF_CNEC}.
    \item \textbf{BSNLP 2019} shared task dataset \cite{piskorski-etal-2019-second} that consists of 196 train and 302 test sentences. We further split the test dataset into development and test parts resulting in development and test datasets of sizes 149 and 153 sentences, respectively. Additionally, we convert the original dataset into the same format as the \textit{CNEC}, extracting entity classes only.
\end{enumerate}

Independently on the dataset, we pre-process the sentences so that the maximum length of an example is 128 sub-word tokens. If the maximum length is exceeded, the residual part is used to create another data point. On the contrary, if the maximum length is not reached, the sentence is padded (padding is inserted at the end of the sentence). It is worth mentioning that exceeding the maximum length of a sentence occurs only for 44 times on the \textit{CNEC}, which is negligible. On the other hand, on the \textit{BSNLP 2019}, the length of the sentences differs a lot, and the maximum length is exceeded for a significant portion of the data.  However, our experiments show that increasing the maximum sequence length does not improve the resulting F1 score. The architecture of the model follows the token classification settings described in Section \ref{sec:tasks}. See Appendix \ref{token_classification_models} for more details about the model and hyper-parameters. 




\subsubsection{Results}
As an evaluation metric, we use F1 score computed on the entity level, while ignoring "O" (empty) class. The results, stated with 95\% confidence intervals, are summarized in Table \ref{tab:NER}.

\begin{table}[ht!]
\centering
\begin{adjustbox}{width=0.8\linewidth,center}
\begin{tabular}{lcc}
\toprule
         & CNEC                        & BSNLP 2019                   \\
\midrule
mBERT         & $86.23 \pm 0.21$ & $84.01 \pm 1.25$                      \\
 SlavicBERT       & $\textbf{86.57} \pm \textbf{0.12}$ & $\textbf{86.70} \pm \textbf{0.37}$    \\
ALBERT-r & $ 34.64 \pm 0.34$                  & $ 19.77 \pm 0.94$                     \\
Czert-A       & $72.95 \pm 0.23$                   & $48.86 \pm 0.61$                      \\
Czert-B       & $86.27 \pm 0.12$                   & $\textbf{86.73} \pm \textbf{0.34}$                      \\ 
\midrule
SoTA& $81.77$ \textsuperscript{b}  &\textbf{ 93.9} \textsuperscript{a} \\
\bottomrule

\end{tabular}
\end{adjustbox}
\caption{Comparison of F1 score achieved using pre-trained Czert-A, Czert-B, mBERT, SlavicBERT and randomly initialised ALBERT on NER task. \textsuperscript{b}Taken from \citet{konopik18lda}
\textsuperscript{a}Taken from \cite{arkhipov2019tuning-SlavicBert}.
}
\label{tab:NER}
\end{table}

\subsection{Morphological Tagging}\label{sec:T_POSTAG}
To evaluate our model on a morphological tagging task, we utilize four Universal Dependencies treebanks. These are namely: Prague Dependency Treebank 3.0 (PDT) \cite{BeHaPragueDependency2013}, Czech Academic Corpus 2.0 \cite{CAC}, Czech Legal Text Treebank 2.0 \cite{CLTT} and FicTree \cite{FicTree}. Together they comprise 103 143 train, 11 326 development and 12 216 test examples. Furthermore, we also perform our experiments on the PDT only to compare our model to the current SoTA. The PDT dataset then comprises 68 627 train, 9 285 dev and 10 163 test examples. The original datasets come as CoNLL files which we converted to a simplified format as in the case of the \textit{CNEC} dataset (section \ref{sec:T_NER}). During this pre-processing step, we extracted only \textit{UPOS} tags, which we use as labels. 
The architecture of the model follows the token classification settings described in Section \ref{sec:tasks}. The number of output neurons is set to the number of possible \textit{UPOS} tags.
See \ref{app:morphtag-hyper}, for more details about the hyper-parameters and training process.



\subsubsection{Results}
Table \ref{tab:UD} shows the achieved results with 95\% confidence intervals. Results are stated in F1 score computed on a token level, ignoring the "O" (empty) class. As the table shows, our model \textit{Czert-B} outperforms the other models on both datasets. Moreover, we outperformed the current SoTA \cite{straka2019czech} as well.

\begin{table}[ht!]
\centering
\begin{adjustbox}{width=1\linewidth,center}

\begin{tabular}{lcc}

\toprule
              &                      Universal Dependencies & PDT \\
\midrule
mBERT         & $99.176 \pm 0.006$ &    $99.301 \pm 0.005$                    \\
SlavicBERT       & $99.211 \pm 0.008$ & $99.318 \pm 0.008$                          \\
ALBERT-r & $ 96.590 \pm 0.096$  & $96.410 \pm 0.060$                       \\
Czert-A       & $98.713 \pm 0.008$   & $97.028 \pm 0.023$                       \\
Czert-B      & $\textbf{99.300} \pm \textbf{0.009}$ &  $\textbf{99.410} \pm \textbf{0.006}$       \\
\midrule
SoTA  &    & 99.34\textsuperscript{a}      \\
\bottomrule
\end{tabular}

\end{adjustbox}
\caption{Comparison of F1 score achieved using pre-trained Czert-A, Czert-B, mBERT, SlavicBERT and randomly initialised ALBERT on morphological tagging task.
\label{tab:UD}
\textsuperscript{a}Result is taken from \cite{straka2019czech}.
}
\end{table}

\subsection{Semantic Role Labelling}\label{sec:T_SRL}


In semantic role labeling we are looking for shallow semantic structure so the task can be formalized as classification of roles arguments of the predicates in the sentence. Therefore, a single example to be classified is the pair of predicate and argument where the predicate is a single word, and the argument is either word or a phrase. We are classifying the role of the argument towards the predicate. Our input representation is inspired by \cite{shi2019simple}. We first tokenize the sentence with WordPiece. Then we feed the sentence into the network followed by the \texttt{[CLS]} token and the predicate token(s). Note that the predicate tokens have the same positional IDs as their occurrence in the sentence, but different segment ids. This way the predicate at the end of the sequence differs from its in-sentence representation only in segment embedding, so it contains all the information to encode the in-sentence context but it can be easily distinguished from other tokens by the segment embedding.

\subsubsection{Results}
We evaluate Semantic role labeling for the Czech language on the CoNLL 2009 dataset. The results are shown in Table \ref{tab:SRL}; the \textit{dep-based} column denotes the result achieved by \citet{zhao2009multilingual}. In \textit{gold-dep}, we replicated their system but evaluated it with gold-standard dependency trees. Syntax-based F1 metric\footnote{Oficial evaluation metric of CoNLL 2009 task.} is computed on whole subtrees of dependency trees. To compute this for span based model, we need to project labels on dependency trees. We did not optimize this projection in any way\footnote{Because we do not want to add information from gold dependency tree annotations.}. We just removed \textit{B-} and \textit{I-} prefixes, we copied the dependency annotation and ran the CoNLL 20009 evaluation script.

\begin{table}[]
\centering
\begin{adjustbox}{width=0.8\linewidth,center}

\begin{tabular}{lcc}
\toprule
\textbf{} &SPAN    & SYNTAX        \\
\midrule
mBERT         & $78.55\pm 0.11$                  & $90.23\pm0.22$                   \\ 
SlavicBERT        & $79.33\pm0.08$ & $90.49\pm0.04$ \\ 
ALBERT-r & $51.37\pm 0.42$                  & $80.75\pm0.13$                   \\ 
Czert-A       & $76.63\pm0.13$                   & $89.94\pm0.05$                   \\ 
Czert-B       & $\textbf{81.86}\pm\textbf{0.10}$ & $\textbf{91.46}\pm\textbf{0.06}$ \\ 
\midrule
dep-based     & -                                 & $85.19$                           \\ 
gold-dep      & -                                 & $89.52$                           \\ 
\bottomrule
\end{tabular}
\end{adjustbox}
\caption{SRL results --  dep columns are evaluate with labelled F1 from CoNLL 2009 evaluation script, other columns are evaluated with span F1 score same as it was used for NER evaluation.}
\label{tab:SRL}
\end{table}

As we can see from the table, \textit{Czert-B} and \textit{SlavicBERT} significantly outperform the other models and they even outperform tree-based approach with gold-standard trees. \textit{Czert-B} and \textit{SlavicBERT} performance are very similar in this task.

\subsection{Sentiment Classification}\label{sec:T_SENT}
Sentiment Classification (SC) task \cite{liu2012sentiment} also called \textit{Polarity Detection}, is a classification task where the goal is to assign a sentiment polarity of a given text. The \textit{positive, negative} and \textit{neutral} classes are usually used as the sentiment polarity labels.
We perform the evaluation on two Czech sentiment classification datasets from \citet{habernal-etal-2013-sentiment}, consisting of (1) Facebook posts and (2) movie reviews.

\par The Facebook dataset (\textit{FB}) contains 10K users’ posts taken from nine Czech Facebook pages annotated with three\footnote{The dataset contains also 248 samples with a fourth class \textit{bipolar} which we do not use.} classes. 


We split the datasets into train, development and test parts
with class distribution that follows the original datasets.

We fine-tune the models separately for each dataset. The architecture of the model follows the sequence pair classification setting described in Section \ref{sec:tasks}. The number of output neurons is set to the number of sentiment polarity classes. \emph{Softmax} normalization is applied to the output layer. We employ \emph{Cross-entropy} loss. See \ref{app:sentiment}, for more details about hyper-parameters.

\subsubsection{Results}
We fine-tune the models (including the baselines) to achieve the best F1 score on the development data. Then, we use the best model settings to train a model on the train and development data. Then, this model is evaluated on the test data and results are reported in Table \ref{tab:sentiment} along with the initial learning rate and the number of epochs used for training. We repeat each experiment six times, and we report the average F1 score along with the 95\% confidence interval.

\begin{table}[ht!]
\begin{adjustbox}{width=\columnwidth,center}
\begin{tabular}{lcc}
\toprule
                  &  FB                                & CSFD           \\ 
\midrule
mBERT         & $71.72 \pm 0.91$ \footnotesize{(2e-5 / 6)}               & $82.80 \pm 0.14$ \footnotesize{(2e-6 / 13)}                      \\
SlavicBERT        & $73.87 \pm 0.50$ \footnotesize{(2e-5 / 3)}                 & $82.51 \pm 0.14$ \footnotesize{(2e-6 / 12)}                  \\
ALBERT-r & $59.50 \pm 0.47$ \footnotesize{(2e-6 / 14)}                  & $75.40 \pm 0.18$ \footnotesize{(2e-6 / 13)}                   \\
Czert-A       & $72.47 \pm 0.72$  \footnotesize{(2e-5 / 8)}                 & $79.58 \pm 0.46$ \footnotesize{(2e-6 / 8)}                  \\
Czert-B       & $\textbf{76.55} \pm \textbf{0.14}$ \footnotesize{(2e-6 / 12)} & $\textbf{84.79} \pm \textbf{0.26}$ \footnotesize{(2e-5 / 12)} \\ \midrule
SoTA       &  69.4\textsuperscript{a}    &  $80.5 \pm 0.16$\textsuperscript{b} \\
\bottomrule
\end{tabular}
\end{adjustbox}
\caption{Average F1 results for the Sentiment Classification task. The numbers in the brackets denote the initial learning rate and number of epochs, respectively, for training of the corresponding model.
The state-of-the-art results \textsuperscript{a} are taken from \cite{habernal-etal-2013-sentiment} and  \textsuperscript{b} \cite{sido2019curriculum}.}
\label{tab:sentiment}
\end{table}

We can see that our Czert-B model outperforms all other models by a large margin on both datasets. We also observe (not shown in the results) for all models that lower initial learning rates (i.e., 2e-6 and 2e-5) lead to more stable fine-tuning than using the initial learning rate of 2.5e-5 which tends to overfit more often as we found out when repeating the experiments. Results for the FB dataset have relatively wide confidence intervals (except for the Czert-B), we believe that it is caused by the small size of the dataset.

\subsection{Multi-label Document Classification}\label{sec:T_MULTI}

Multi-label Document Classification is a variant of classification problem where multiple labels can be assigned to each document. In this problem, there is no constraint on how many of the labels can be assigned to a given document.

We work with the \textit{Czech Text Document Corpus v 1.0} \cite{Kral13CORES} to fine-tune and evaluate the models. The Czech News Agency provided almost 12 thousands of documents that formed the basis of this dataset. The agency journalists assign 60 categories (tags) to the documents as a part of their daily work. Following the approach from \cite{kral_cicling16a}, we use only 37 most frequent categories for evaluation. More statistics are available in the paper.

\subsubsection{Model Description and Fine-tuning}
For \textit{multi-label classification of documents} (MLC), 
we follow the sequence classification setting described in Section \ref{sec:tasks}. 
The output layer is activated by the \textit{sigmoid} function. The loss is the \emph{Binary Cross-entropy} function. In the context of this task, documents are regarded as sentences trimmed to the maximum sequence length in tokens set to 512. We chose to pick the first N tokens in each document as our trimming strategy. 

We run twenty 10-epoch-long training phases for each model and average the results. See \ref{app:mlc} for more details.

We use both standard \emph{F1} and the \emph{AUROC} \cite{Melo2013} evaluation metrics. AUROC represents the overall ability of MLC models to distinguish between different classes without being biased by any constant threshold value. We use 95\% confidence interval. We present the results in Table \ref{tab:mlc}.

\begin{table}[]
\centering
\begin{adjustbox}{width=0.8\linewidth,center}
\begin{tabular}{lcc}
\toprule
         & \multicolumn{2}{c}{CTDC-1}                 \\ \cline{2-3} 
         & AUROC                              & F1 \\
\toprule
mBERT      & $97.62 \pm 0.08$                   &  $83.04 \pm 0.16$  \\
SlavicBERT & $97.80 \pm 0.06$                   &  $84.08 \pm 0.14$   \\
ALBERT-r   & $94.35 \pm 0.13$                   &  $72.44 \pm 0.22$  \\
Czert-A    & $97.49 \pm 0.07$                    &  $82.27 \pm 0.17$  \\
Czert-B    & $\textbf{98.00} \pm \textbf{0.04}$ &  $\textbf{85.06} \pm \textbf{0.11}$  \\
\midrule
SoTA       &    --            &  \textbf{84.7}\textsuperscript{*}  \\
\bottomrule
\end{tabular}
\end{adjustbox}
\caption{
Results for Multi-label Document Classification on Czech Text Document Corpus v 1.0 dataset -- AUROC and F1 measures. SoTA taken from \cite{kral_cicling16a}. 
}
\label{tab:mlc}
\end{table}

\subsection{Semantic Text Similarity}\label{sec:T_STS}
We evaluate our model on semantic text similarity task on two different datasets. 
\begin{enumerate}
    \item \textit{STS-SVOB} \cite{svoboda2018czech} contains two datasets: images descriptions (550 train and 300 test samples); and headlines (375 train and 200 test samples).
We use the raw variant without any lemmatization or stemming. 

    \item \textit{STS-CNA}  was created during our experiments with this new model in cooperation with Czech News Agency and Charles University. STS-CNA contains s 138,556 hand-annotated sentence pairs \cite{sido2021CzechNewsForSts}.
    
    
\end{enumerate}


\subsubsection{Model Description and Fine-tuning} 
The architecture of the model follows the sequence pair classification setting described in Section \ref{sec:tasks}. The number of output neurons is set to 1, and no activation function is applied to the output layer. We employ the \emph{Mean Squared Error} loss.

We tried to keep hyper-parameters as close as possible between all experiments; however, we were forced to change them slightly in case of Czert-A and ALBERT-r.
 Also, the datasets have different nature; thus, we use different sets of hyper-parameters for each dataset.
 See \ref{app:sts}




We run ten experiments for each configuration and use 95\% confidence interval. The tables Table \ref{tab:sts1} and Table \ref{tab:sts2} summarize the results. Table \ref{tab:sts1} shows that Czert-B model significantly outperforms the SoTA on SVOB-IMG dataset. In the SVOB-HL dataset, the models perform in par. We believe that the draw can be caused by reaching the annotation accuracy limit of this dataset.  

\begin{table}[]
\begin{adjustbox}{width=\columnwidth,center}
\centering
\begin{tabular}{lccc}

\toprule
                                          &                  STS-CNA      &             SVOB-IMG     &     SVOB-HL    \\
                                          \midrule
mBERT  & $90.93\pm0.34$                   & $79.37 \pm 0.49$                     & $78.83 \pm 0.30$                   \\
SlavicBERT & $91.38\pm0.29$                   & $79.90 \pm 0.81 $                    & $77.00 \pm 0.31$                   \\
ALBERT-r   & $43.18\pm0.13$                  & $15.74 \pm 2.99$                    & $33.95 \pm 1.81$                   \\
Czert-A         & $88.72\pm0.25$                   & $79.444 \pm 0.34$                    & $ 75.09\pm 0.81$                   \\
Czert-B         & $\textbf{91.89}\pm\textbf{0.12}$ & $\textbf{83.74} \pm \textbf{0.40}$ & $\textbf{79.83} \pm \textbf{0.47}$\\
\midrule
\textit{SoTA}\textsuperscript{*} & -- & \textit{78.87} & \textbf{79.99}  \\
\bottomrule

\end{tabular}
\end{adjustbox}
\caption{Pearson correlation (95\% conf. from ten experiments). \textsuperscript{*}Taken from
\label{tab:sts1}
\citet{svoboda2018czech}}
\end{table}

\begin{table}[]

\begin{adjustbox}{width=\columnwidth,center}
\begin{tabular}{lccc}
\toprule
              &                    CNA       &                    SVOB-IMG    &  STS-SVOB-HL                        \\
              \midrule
mBERT         & $87.88\pm0.08$                   & $78.83 \pm 0.36$                   & $\textbf{78.83} \pm \textbf{0.37}$ \\
SlavicBERT        & $88.97\pm0.09$                   & $79.66 \pm0.73 $                   & $76.03 \pm 0.42$                   \\
ALBERT-r & $33.32\pm0.11$                   & $15.15 \pm 3.07$                   & $32.25 \pm 2.05$                   \\
Czert-A       & $85.85\pm0.16$                   & $78.72 \pm 0.38$                   & $73.86 \pm 0.72$                   \\
Czert-B       & $\textbf{89.29}\pm\textbf{0.17}$ & $\textbf{83.20} \pm \textbf{0.39}$ & $\textbf{78.69} \pm \textbf{0.59}$\\
\bottomrule
\end{tabular}
\end{adjustbox}
\caption{Spearman correlation (95\% conf. from ten experiments)}
\label{tab:sts2}
\end{table}

We also observe a more stable and robust training on extremely small datasets; both Czert models are less prone to over-fitting than other tested models.

\section{Discussion}

\begin{table*}[ht!]
\catcode`\-=12
\begin{adjustbox}{width=\linewidth,center}
\begin{tabular}{lcccccccccccccccc}
\hline
                                         & \multicolumn{2}{c}{NER}                                                                                    &  & \multicolumn{2}{c}{MoT}                   &  & SRL                                                                   &  & \multicolumn{2}{c}{SENTIMENT}                                                                                                                 &  & MULTI-CLASS                                                           &  & \multicolumn{3}{c}{STS}                                                                                                                                                                                       \\ \cline{2-3} \cline{5-6} \cline{8-8} \cline{10-11} \cline{13-13} \cline{15-17} 
                                         & CENEC                                                                 & BSNLP                              &  & UNIV. DEP.                          & PDT &  & CoNLL-09                                                              &  & FB                                                                    & CSFD                                                                  &  & CTDC-1                                                                &  & CNA                                                                   & SVOB-IMG                                                               & SVOB-HL                                                      \\ \hline
mBERT                                    & 86.23                                                                 & 84.01                              &  & 99.176                              & 99.301 &  & 90.23                                                                 &  & 71.72                                                                 & 81.35                                                                 &  & 83.04                                                                 &  & 90.93                                                                 & 79.37                                                                  & 78.83                                                        \\
SlavicBERT                               & \underline{\textbf{86.57}}$\dagger$ & \underline{86.70} &  & 99.211                              & 99.318 &  & 90.49                                                                 &  & 73.87                                                                 & 81.55                                                                 &  & 84.08                                                                 &  & 91.38                                                                 & 79.90                                                                  & 77.00                                                        \\
ALBERT-r                                 & 34.64                                                                 & 19.77                              &  & 96.590                              & 96.410 &  & 80.75                                                                 &  & 59.50                                                                 & 70.33                                                                 &  & 72.44                                                                 &  & 43.18                                                                 & 15.73                                                                  & 33.95                                                        \\
Czert-A                                  & 72.95                                                                 & 48.86                              &  & 98.713                              & 97.028 &  & 89.94                                                                 &  & 72.47                                                                 & 79.73                                                                 &  & 82.27                                                                 &  & 88.72                                                                 & 79.44                                                                  & 75.09                                                        \\
Czert-B                                  & 86.27                                                                 & \underline{86.73} &  & \underline{\textbf{99.300}}$\dagger$ & \underline{\textbf{99.410}}$\dagger$ &  & \underline{\textbf{91.46}}$\dagger$ &  & \underline{\textbf{76.55}}$\dagger$ & \underline{\textbf{84.79}}$\dagger$ &  & \underline{\textbf{85.06}}$\dagger$ &  & \underline{\textbf{91.89}}$\dagger$ & \underline{\textbf{83.74}} $\dagger$ & \underline{\textbf{79.83}} \\ \hline
SoTA\textsuperscript{*} & 81.77                                                                 & \textbf{93.9}     &  & --                                 & 99.34 &  & 89.52                                                                 &  & 69.4                                                                  & 80.5                                                                  &  & 84.7                                                                  &  & --                                                                    & 78.87                                                                  & \textbf{79.99}                              \\ \hline
\end{tabular}

\end{adjustbox}
\caption{Summary of our results. The bold results denote the current SoTA results. The underlined results are the best result achieved directly by fine-tuning the BERT-like models. Values with the $\dagger$ symbols are the new SoTA results that we established in this paper. \textsuperscript{*}Results are taken from original papers.}\label{tab:overall}
\end{table*}

We summarize the overall results of all evaluated tasks in Table \ref{tab:overall}. The first three columns contain the token classification tasks, the next two columns show results for sequence classification tasks, and the last column belongs to sequence pair classification task. We can observe that Czert-B model excels at the sequence and sequence pair classification tasks. In these tasks, Czert-B outperforms other pre-trained models by a large margin. We believe that the likely cause for such results lay in the amount of Czech data we use to train Czert models. mBERT and SlavicBERT use only Czech Wikipedia, but we work with almost 50 times larger data in terms of sentence count.
For most of the token classification tasks, Czert-B performs similarly to other pre-trained models except for SRL, where Czert-B outperformed other models by a large margin.
We establish a new state of the art on \textbf{NER} with the SlavicBERT model on the CNEC dataset. The performance increase is a major one. We increase the F1 measure by 5\%.
Also, we achieve similar results with SlavicBERT and Czert-B on BSNLP dataset. 

We also outperformed other BERT-like models with Czert-B in \textbf{MoT}, and surpass the SoTA. 


We accomplish outstanding performance and increase the SoTA in two other tasks: sentiment classification (\textbf{SC}) and semantic text similarity (\textbf{STS}). The increase is of $\sim$5\% and $\sim$3\% in both sentiment datasets and of $\sim$5\% in one of the semantic similarity datasets. We also overcame SoTA in \textbf{MLC}.



\section{Conclusion}
In this work, we present two monolingual BERT-like models (BERT and ALBERT) for the Czech language. We train the models with the original MLM task and with a slightly modified NSP task. We thoroughly evaluate our models on six common tasks, and we compare them with other multilingual models. We include task-specific state-of-the-art models in our comparison. We outperform multilingual models with our newly trained Czert-B model on 9 out of 11 datasets. In addition, we establish the new state-of-the-art results on 9 datasets\footnote{The results in Table \ref{tab:overall} with the $\dagger$ symbol.}. The results show the strong performance of the Czert-B model on STS, MLC, SC, SRL, and MoT tasks. As our paper confirms and as is shown in similar works, monolingual Transformer-based language models often overcome the multilingual ones.


\par Our models are publicly available for research purposes at our website and in the hugging face repository\footnote{\url{https://huggingface.co/UWB-AIR}}.

\section*{Acknowledgement}
This work has been partly supported by ERDF ”Research and Development of Intelligent Components of Advanced Technologies for the Pilsen Metropolitan Area (InteCom)” (no.: CZ.02.1.01/0.0/0.0/17 048/0007267); and by Grant No. SGS-2019-018 Processing of heterogeneous data and its specialized applications. Computational resources were supplied by the project "e-Infrastruktura CZ" (e-INFRA LM2018140) provided within the program Projects of Large Research, Development and Innovations Infrastructures.

\bibliography{naacl2021}

\begin{thebibliography}{48}
\expandafter\ifx\csname natexlab\endcsname\relax\def\natexlab#1{#1}\fi

\bibitem[{Abadi et~al.(2015)Abadi, Agarwal, Barham, Brevdo, Chen, Citro,
  Corrado, Davis, Dean, Devin, Ghemawat, Goodfellow, Harp, Irving, Isard, Jia,
  Jozefowicz, Kaiser, Kudlur, Levenberg, Man\'{e}, Monga, Moore, Murray, Olah,
  Schuster, Shlens, Steiner, Sutskever, Talwar, Tucker, Vanhoucke, Vasudevan,
  Vi\'{e}gas, Vinyals, Warden, Wattenberg, Wicke, Yu, and
  Zheng}]{tensorflow2015-whitepaper}
Mart\'{\i}n Abadi, Ashish Agarwal, Paul Barham, Eugene Brevdo, Zhifeng Chen,
  Craig Citro, Greg~S. Corrado, Andy Davis, Jeffrey Dean, Matthieu Devin,
  Sanjay Ghemawat, Ian Goodfellow, Andrew Harp, Geoffrey Irving, Michael Isard,
  Yangqing Jia, Rafal Jozefowicz, Lukasz Kaiser, Manjunath Kudlur, Josh
  Levenberg, Dandelion Man\'{e}, Rajat Monga, Sherry Moore, Derek Murray, Chris
  Olah, Mike Schuster, Jonathon Shlens, Benoit Steiner, Ilya Sutskever, Kunal
  Talwar, Paul Tucker, Vincent Vanhoucke, Vijay Vasudevan, Fernanda Vi\'{e}gas,
  Oriol Vinyals, Pete Warden, Martin Wattenberg, Martin Wicke, Yuan Yu, and
  Xiaoqiang Zheng. 2015.
\newblock \href {https://www.tensorflow.org/} {{TensorFlow}: Large-scale
  machine learning on heterogeneous systems}.
\newblock Software available from tensorflow.org.

\bibitem[{Arkhipov et~al.(2019)Arkhipov, Trofimova, Kuratov, and
  Sorokin}]{arkhipov2019tuning-SlavicBert}
Mikhail Arkhipov, Maria Trofimova, Yuri Kuratov, and Alexey Sorokin. 2019.
\newblock \href {https://doi.org/10.18653/v1/W19-3712} {Tuning multilingual
  transformers for language-specific named entity recognition}.
\newblock In \emph{Proceedings of the 7th Workshop on Balto-Slavic Natural
  Language Processing}, pages 89--93, Florence, Italy. Association for
  Computational Linguistics.

\bibitem[{Bej{\v{c}}ek et~al.(2013)Bej{\v{c}}ek, Haji{\v{c}}ov{\'{a}},
  Haji{\v{c}}, J{\'{i}}nov{\'{a}}, Kettnerov{\'{a}},
  Kol{\'{a}}{\v{r}}ov{\'{a}}, Mikulov{\'{a}}, M{\'{i}}rovsk{\'{y}}, Nedoluzhko,
  Panevov{\'{a}}, Pol{\'{a}}kov{\'{a}}, {\v{S}}ev{\v{c}}{\'{i}}kov{\'{a}},
  {\v{S}}t{\v{e}}p{\'{a}}nek, and
  Zik{\'{a}}nov{\'{a}}}]{BeHaPragueDependency2013}
Eduard Bej{\v{c}}ek, Eva Haji{\v{c}}ov{\'{a}}, Jan Haji{\v{c}}, Pavl{\'{i}}na
  J{\'{i}}nov{\'{a}}, V{\'{a}}clava Kettnerov{\'{a}}, Veronika
  Kol{\'{a}}{\v{r}}ov{\'{a}}, Marie Mikulov{\'{a}}, Ji{\v{r}}{\'{i}}
  M{\'{i}}rovsk{\'{y}}, Anna Nedoluzhko, Jarmila Panevov{\'{a}}, Lucie
  Pol{\'{a}}kov{\'{a}}, Magda {\v{S}}ev{\v{c}}{\'{i}}kov{\'{a}}, Jan
  {\v{S}}t{\v{e}}p{\'{a}}nek, and {\v{S}}{\'{a}}rka Zik{\'{a}}nov{\'{a}}. 2013.
\newblock Prague dependency treebank 3.0.

\bibitem[{Bojanowski et~al.(2017)Bojanowski, Grave, Joulin, and
  Mikolov}]{bojanowski2017enriching}
Piotr Bojanowski, Edouard Grave, Armand Joulin, and Tomas Mikolov. 2017.
\newblock Enriching word vectors with subword information.
\newblock \emph{Transactions of the Association for Computational Linguistics},
  5:135--146.

\bibitem[{Conneau et~al.(2020)Conneau, Khandelwal, Goyal, Chaudhary, Wenzek,
  Guzm{\'a}n, Grave, Ott, Zettlemoyer, and
  Stoyanov}]{conneau-etal-2020-unsupervised-XLM-Roberta}
Alexis Conneau, Kartikay Khandelwal, Naman Goyal, Vishrav Chaudhary, Guillaume
  Wenzek, Francisco Guzm{\'a}n, Edouard Grave, Myle Ott, Luke Zettlemoyer, and
  Veselin Stoyanov. 2020.
\newblock \href {https://doi.org/10.18653/v1/2020.acl-main.747} {Unsupervised
  cross-lingual representation learning at scale}.
\newblock In \emph{Proceedings of the 58th Annual Meeting of the Association
  for Computational Linguistics}, pages 8440--8451, Online. Association for
  Computational Linguistics.

\bibitem[{Conneau and Lample(2019)}]{nips-2019-XLM}
Alexis Conneau and Guillaume Lample. 2019.
\newblock \href
  {https://proceedings.neurips.cc/paper/2019/file/c04c19c2c2474dbf5f7ac4372c5b9af1-Paper.pdf}
  {Cross-lingual language model pretraining}.
\newblock In \emph{Advances in Neural Information Processing Systems},
  volume~32, pages 7059--7069. Curran Associates, Inc.

\bibitem[{Devlin et~al.(2018)Devlin, Chang, Lee, and Toutanova}]{bertpaper}
Jacob Devlin, Ming{-}Wei Chang, Kenton Lee, and Kristina Toutanova. 2018.
\newblock \href {http://arxiv.org/abs/1810.04805} {{BERT:} pre-training of deep
  bidirectional transformers for language understanding}.
\newblock \emph{CoRR}, abs/1810.04805.

\bibitem[{Dumitrescu et~al.(2020)Dumitrescu, Avram, and
  Pyysalo}]{dumitrescu2020birth-romanianBERT}
Stefan Dumitrescu, Andrei-Marius Avram, and Sampo Pyysalo. 2020.
\newblock \href {https://www.aclweb.org/anthology/2020.findings-emnlp.387} {The
  birth of {R}omanian {BERT}}.
\newblock In \emph{Proceedings of the 2020 Conference on Empirical Methods in
  Natural Language Processing: Findings}, pages 4324--4328, Online. Association
  for Computational Linguistics.

\bibitem[{Habernal et~al.(2013)Habernal, Pt{\'a}{\v{c}}ek, and
  Steinberger}]{habernal-etal-2013-sentiment}
Ivan Habernal, Tom{\'a}{\v{s}} Pt{\'a}{\v{c}}ek, and Josef Steinberger. 2013.
\newblock \href {https://www.aclweb.org/anthology/W13-1609} {Sentiment analysis
  in {C}zech social media using supervised machine learning}.
\newblock In \emph{Proceedings of the 4th Workshop on Computational Approaches
  to Subjectivity, Sentiment and Social Media Analysis}, pages 65--74, Atlanta,
  Georgia. Association for Computational Linguistics.

\bibitem[{Hn{\'{a}}tkov{\'{a}} et~al.(2017)Hn{\'{a}}tkov{\'{a}}, Jel{\'{i}}nek,
  Kl{\'{i}}mov{\'{a}}, Krop{\'{i}}kov{\'{a}}, Skoumalov{\'{a}}, Zitov{\'{a}},
  and Zeman}]{FicTree}
Milena Hn{\'{a}}tkov{\'{a}}, Tom{\'{a}} Jel{\'{i}}nek, Ivana
  Kl{\'{i}}mov{\'{a}}, Alena Krop{\'{i}}kov{\'{a}}, Hana Skoumalov{\'{a}}, Olga
  Zitov{\'{a}}, and Daniel Zeman. 2017.
\newblock Fictree.

\bibitem[{Howard and Ruder(2018)}]{howard-ruder-2018-ULMFIT}
Jeremy Howard and Sebastian Ruder. 2018.
\newblock \href {https://doi.org/10.18653/v1/P18-1031} {Universal language
  model fine-tuning for text classification}.
\newblock In \emph{Proceedings of the 56th Annual Meeting of the Association
  for Computational Linguistics (Volume 1: Long Papers)}, pages 328--339,
  Melbourne, Australia. Association for Computational Linguistics.

\bibitem[{Hrala and Kr\'al(2013)}]{Kral13CORES}
M.~Hrala and P.~Kr\'al. 2013.
\newblock \href {https://doi.org/10.1007/978-3-319-00969-8\_86} {Evaluation of
  the document classification approaches}.
\newblock In \emph{8th International Conference on Computer Recognition Systems
  (CORES 2013)}, pages 877--885, Milkow, Poland. Springer.

\bibitem[{Kingma and Ba(2014)}]{Kingma-adam}
Diederik~P Kingma and Jimmy Ba. 2014.
\newblock Adam: A method for stochastic optimization.
\newblock \emph{arXiv preprint arXiv:1412.6980}.

\bibitem[{Konkol and Konop\'{i}k(2013)}]{CRF_CNEC}
Michal Konkol and Miloslav Konop\'{i}k. 2013.
\newblock Crf-based czech named entity recognizer and consolidation of czech
  ner research.
\newblock In \emph{Text, Speech and Dialogue}, volume 8082 of \emph{Lecture
  Notes in Computer Science}, pages 153--160. Springer Berlin Heidelberg.

\bibitem[{Konop{\'i}k and Pra{\v{z}}{\'a}k(2018)}]{konopik18lda}
Miloslav Konop{\'i}k and Ond{\v{r}}ej Pra{\v{z}}{\'a}k. 2018.
\newblock \href {https://doi.org/10.1007/978-3-030-00794-2_6} {Lda in
  character-lstm-crf named entity recognition}.
\newblock In \emph{International Conference on Text, Speech, and Dialogue},
  pages 58--66, Cham. Springer International Publishing.

\bibitem[{K{\v r}en et~al.(2016)K{\v r}en, Cvr{\v c}ek, {\v C}apka, {\v
  C}erm{\'a}kov{\'a}, Hn{\'a}tkov{\'a}, Chlumsk{\'a}, Jel{\'{\i}}nek,
  Kov{\'a}{\v r}{\'{\i}}kov{\'a}, Petkevi{\v c}, Proch{\'a}zka, Skoumalov{\'a},
  {\v S}krabal, Trune{\v c}ek, Vond{\v r}i{\v c}ka, and Zasina}]{cs-nat-corpus}
Michal K{\v r}en, V{\'a}clav Cvr{\v c}ek, Tom{\'a}{\v s} {\v C}apka, Anna {\v
  C}erm{\'a}kov{\'a}, Milena Hn{\'a}tkov{\'a}, Lucie Chlumsk{\'a}, Tom{\'a}{\v
  s} Jel{\'{\i}}nek, Dominika Kov{\'a}{\v r}{\'{\i}}kov{\'a}, Vladim{\'{\i}}r
  Petkevi{\v c}, Pavel Proch{\'a}zka, Hana Skoumalov{\'a}, Michal {\v S}krabal,
  Petr Trune{\v c}ek, Pavel Vond{\v r}i{\v c}ka, and Adrian Zasina. 2016.
\newblock \href {http://hdl.handle.net/11234/1-1846} {{SYN} v4: large corpus of
  written czech}.
\newblock {LINDAT}/{CLARIAH}-{CZ} digital library at the Institute of Formal
  and Applied Linguistics ({{\'U}FAL}), Faculty of Mathematics and Physics,
  Charles University.

\bibitem[{K{\v{r}}{\'{i}}{\v{z}} et~al.(2018)K{\v{r}}{\'{i}}{\v{z}},
  Hladk{\'{a}}, and Ure{\v{s}}ov{\'{a}}}]{CLTT}
Vincent K{\v{r}}{\'{i}}{\v{z}}, Barbora Hladk{\'{a}}, and Zde{\v{n}}ka
  Ure{\v{s}}ov{\'{a}}. 2018.
\newblock Czech legal text treebank 2.0.

\bibitem[{Kłeczek(2020)}]{Kleczek2020-polbert}
Dariusz Kłeczek. 2020.
\newblock Polbert: Attacking polish nlp tasks with transformers.
\newblock In \emph{Proceedings of the PolEval 2020 Workshop}. Institute of
  Computer Science, Polish Academy of Sciences.

\bibitem[{Lan et~al.(2020)Lan, Chen, Goodman, Gimpel, Sharma, and
  Soricut}]{lan2019albert}
Zhenzhong Lan, Mingda Chen, Sebastian Goodman, Kevin Gimpel, Piyush Sharma, and
  Radu Soricut. 2020.
\newblock \href {https://openreview.net/forum?id=H1eA7AEtvS} {{ALBERT:} {A}
  lite {BERT} for self-supervised learning of language representations}.
\newblock In \emph{8th International Conference on Learning Representations,
  {ICLR} 2020, Addis Ababa, Ethiopia, April 26-30, 2020}. OpenReview.net.

\bibitem[{Le et~al.(2019)Le, Vial, Frej, Segonne, Coavoux, Lecouteux, Allauzen,
  Crabb{\'e}, Besacier, and Schwab}]{le2019flaubert}
Hang Le, Lo{\"\i}c Vial, Jibril Frej, Vincent Segonne, Maximin Coavoux,
  Benjamin Lecouteux, Alexandre Allauzen, Beno{\^\i}t Crabb{\'e}, Laurent
  Besacier, and Didier Schwab. 2019.
\newblock Flaubert: Unsupervised language model pre-training for french.
\newblock \emph{arXiv preprint arXiv:1912.05372}.

\bibitem[{Lenc and Kr{\'a}l(2018)}]{kral_cicling16a}
Ladislav Lenc and Pavel Kr{\'a}l. 2018.
\newblock \href {https://doi.org/10.1007/978-3-319-75487-1_36} {Deep neural
  networks for {C}zech multi-label document classification}.
\newblock In \emph{Computational Linguistics and Intelligent Text Processing},
  pages 460--471, Cham. Springer International Publishing.

\bibitem[{Liu(2012)}]{liu2012sentiment}
Bing Liu. 2012.
\newblock Sentiment analysis and opinion mining.
\newblock \emph{Synthesis lectures on human language technologies},
  5(1):1--167.

\bibitem[{Liu et~al.()Liu, Ott, Goyal, Du, Joshi, Chen, Levy, Lewis,
  Zettlemoyer, and Stoyanov}]{liu1907roberta}
Y~Liu, M~Ott, N~Goyal, J~Du, M~Joshi, D~Chen, O~Levy, M~Lewis, L~Zettlemoyer,
  and V~Stoyanov.
\newblock Roberta: A robustly optimized bert pretraining approach. arxiv 2019.
\newblock \emph{arXiv preprint arXiv:1907.11692}.

\bibitem[{Martin et~al.(2020)Martin, Muller, Ortiz~Su{\'a}rez, Dupont, Romary,
  de~la Clergerie, Seddah, and Sagot}]{martin2019camembert}
Louis Martin, Benjamin Muller, Pedro~Javier Ortiz~Su{\'a}rez, Yoann Dupont,
  Laurent Romary, {\'E}ric de~la Clergerie, Djam{\'e} Seddah, and Beno{\^\i}t
  Sagot. 2020.
\newblock \href {https://doi.org/10.18653/v1/2020.acl-main.645} {{C}amem{BERT}:
  a tasty {F}rench language model}.
\newblock In \emph{Proceedings of the 58th Annual Meeting of the Association
  for Computational Linguistics}, pages 7203--7219, Online. Association for
  Computational Linguistics.

\bibitem[{McCann et~al.(2017)McCann, Bradbury, Xiong, and Socher}]{cove}
Bryan McCann, James Bradbury, Caiming Xiong, and Richard Socher. 2017.
\newblock \href
  {https://proceedings.neurips.cc/paper/2017/file/20c86a628232a67e7bd46f76fba7ce12-Paper.pdf}
  {Learned in translation: Contextualized word vectors}.
\newblock In \emph{Advances in Neural Information Processing Systems},
  volume~30, pages 6294--6305. Curran Associates, Inc.

\bibitem[{Melo(2013)}]{Melo2013}
Francisco Melo. 2013.
\newblock \href {https://doi.org/10.1007/978-1-4419-9863-7_209} {\emph{Area
  under the ROC Curve}}, pages 38--39. Springer New York, New York, NY.

\bibitem[{Mikolov et~al.(2013)Mikolov, Sutskever, Chen, Corrado, and
  Dean}]{mikolov2013distributedword2vec}
Tomas Mikolov, Ilya Sutskever, Kai Chen, Greg~S Corrado, and Jeff Dean. 2013.
\newblock \href
  {https://proceedings.neurips.cc/paper/2013/file/9aa42b31882ec039965f3c4923ce901b-Paper.pdf}
  {Distributed representations of words and phrases and their
  compositionality}.
\newblock In \emph{Advances in Neural Information Processing Systems},
  volume~26, pages 3111--3119. Curran Associates, Inc.

\bibitem[{Pennington et~al.(2014)Pennington, Socher, and
  Manning}]{pennington-etal-2014-glove}
Jeffrey Pennington, Richard Socher, and Christopher Manning. 2014.
\newblock \href {https://doi.org/10.3115/v1/D14-1162} {{G}lo{V}e: Global
  vectors for word representation}.
\newblock In \emph{Proceedings of the 2014 Conference on Empirical Methods in
  Natural Language Processing ({EMNLP})}, pages 1532--1543, Doha, Qatar.
  Association for Computational Linguistics.

\bibitem[{Peters et~al.(2018)Peters, Neumann, Iyyer, Gardner, Clark, Lee, and
  Zettlemoyer}]{peters-etal-2018-ELMO}
Matthew Peters, Mark Neumann, Mohit Iyyer, Matt Gardner, Christopher Clark,
  Kenton Lee, and Luke Zettlemoyer. 2018.
\newblock \href {https://doi.org/10.18653/v1/N18-1202} {Deep contextualized
  word representations}.
\newblock In \emph{Proceedings of the 2018 Conference of the North {A}merican
  Chapter of the Association for Computational Linguistics: Human Language
  Technologies, Volume 1 (Long Papers)}, pages 2227--2237, New Orleans,
  Louisiana. Association for Computational Linguistics.

\bibitem[{Piskorski et~al.(2019)Piskorski, Laskova, Marci{\'n}czuk, Pivovarova,
  P{\v{r}}ib{\'a}{\v{n}}, Steinberger, and
  Yangarber}]{piskorski-etal-2019-second}
Jakub Piskorski, Laska Laskova, Micha{\l} Marci{\'n}czuk, Lidia Pivovarova,
  Pavel P{\v{r}}ib{\'a}{\v{n}}, Josef Steinberger, and Roman Yangarber. 2019.
\newblock \href {https://www.aclweb.org/anthology/W19-3709} {The second
  cross-lingual challenge on recognition, normalization, classification, and
  linking of named entities across {S}lavic languages}.
\newblock In \emph{Proceedings of the 7th Workshop on Balto-Slavic Natural
  Language Processing}, pages 63--74, Florence, Italy. Association for
  Computational Linguistics.

\bibitem[{Raffel et~al.(2019)Raffel, Shazeer, Roberts, Lee, Narang, Matena,
  Zhou, Li, and Liu}]{raffel2019exploring}
Colin Raffel, Noam Shazeer, Adam Roberts, Katherine Lee, Sharan Narang, Michael
  Matena, Yanqi Zhou, Wei Li, and Peter~J Liu. 2019.
\newblock Exploring the limits of transfer learning with a unified text-to-text
  transformer.
\newblock \emph{arXiv preprint arXiv:1910.10683}.

\bibitem[{Safaya et~al.(2020)Safaya, Abdullatif, and
  Yuret}]{safaya2020kuisail-arabicbert}
Ali Safaya, Moutasem Abdullatif, and Deniz Yuret. 2020.
\newblock \href {http://arxiv.org/abs/2007.13184} {Kuisail at semeval-2020 task
  12: Bert-cnn for offensive speech identification in social media}.

\bibitem[{Sanh et~al.(2019)Sanh, Debut, Chaumond, and
  Wolf}]{sanh2019distilbert}
Victor Sanh, Lysandre Debut, Julien Chaumond, and Thomas Wolf. 2019.
\newblock Distilbert, a distilled version of bert: smaller, faster, cheaper and
  lighter.
\newblock \emph{arXiv preprint arXiv:1910.01108}.

\bibitem[{Schweter(2020)}]{stefan-schweter-BERTurk}
Stefan Schweter. 2020.
\newblock \href {https://doi.org/10.5281/zenodo.3770924} {Berturk - bert models
  for turkish}.

\bibitem[{Sergeev and Balso(2018)}]{sergeev2018horovod}
Alexander Sergeev and Mike~Del Balso. 2018.
\newblock Horovod: fast and easy distributed deep learning in {TensorFlow}.
\newblock \emph{arXiv preprint arXiv:1802.05799}.

\bibitem[{{\v{S}}ev{\v{c}}{\'{\i}}kov{\'{a}}
  et~al.(2007){\v{S}}ev{\v{c}}{\'{\i}}kov{\'{a}}, {\v{Z}}abokrtsk{\'{y}}, and
  Kr{\r{u}}za}]{SevcikovaEtAl2007CNEC}
Magda {\v{S}}ev{\v{c}}{\'{\i}}kov{\'{a}}, Zden{\v{e}}k {\v{Z}}abokrtsk{\'{y}},
  and Old{\v{r}}ich Kr{\r{u}}za. 2007.
\newblock Named entities in czech: Annotating data and developing {NE} tagger.
\newblock In \emph{Lecture Notes in Artificial Intelligence, Proceedings of the
  10th International Conference on Text, Speech and Dialogue}, volume 4629 of
  \emph{Lecture Notes in Computer Science}, pages 188--195, Berlin /
  Heidelberg. Springer.

\bibitem[{Shi and Lin(2019)}]{shi2019simple}
Peng Shi and Jimmy Lin. 2019.
\newblock Simple bert models for relation extraction and semantic role
  labeling.
\newblock \emph{arXiv preprint arXiv:1904.05255}.

\bibitem[{Sido and Konop{\'\i}k(2019)}]{sido2019curriculum}
Jakub Sido and Miloslav Konop{\'\i}k. 2019.
\newblock Curriculum learning in sentiment analysis.
\newblock In \emph{International Conference on Speech and Computer}, pages
  444--450. Springer.

\bibitem[{Sido et~al.(2021)Sido, Seják, Pražák, Konopík, and
  Moravec}]{sido2021CzechNewsForSts}
Jakub Sido, Michal Seják, Ondřej Pražák, Miloslav Konopík, and Václav
  Moravec. 2021.
\newblock \href {http://arxiv.org/abs/2108.08708} {Czech news dataset for
  semanic textual similarity}.
\newblock \emph{arXiv preprint arXiv:2108.08708}.

\bibitem[{Straka et~al.(2019)Straka, Strakov{\'a}, and
  Haji{\v{c}}}]{straka2019czech}
Milan Straka, Jana Strakov{\'a}, and Jan Haji{\v{c}}. 2019.
\newblock Czech text processing with contextual embeddings: Pos tagging,
  lemmatization, parsing and ner.
\newblock In \emph{International Conference on Text, Speech, and Dialogue},
  pages 137--150. Springer.

\bibitem[{Svoboda and Brychc{\'{\i}}n(2018)}]{svoboda2018czech}
Luk{\'{a}}s Svoboda and Tom{\'{a}}s Brychc{\'{\i}}n. 2018.
\newblock \href {https://doi.org/10.1007/978-3-030-00794-2\_23} {Czech dataset
  for semantic textual similarity}.
\newblock In \emph{Text, Speech, and Dialogue - 21st International Conference,
  {TSD} 2018, Brno, Czech Republic, September 11-14, 2018, Proceedings}, volume
  11107 of \emph{Lecture Notes in Computer Science}, pages 213--221. Springer.

\bibitem[{Vaswani et~al.(2017)Vaswani, Shazeer, Parmar, Uszkoreit, Jones,
  Gomez, Kaiser, and Polosukhin}]{attention-all-transformer}
Ashish Vaswani, Noam Shazeer, Niki Parmar, Jakob Uszkoreit, Llion Jones,
  Aidan~N Gomez, \L~ukasz Kaiser, and Illia Polosukhin. 2017.
\newblock \href
  {https://proceedings.neurips.cc/paper/2017/file/3f5ee243547dee91fbd053c1c4a845aa-Paper.pdf}
  {Attention is all you need}.
\newblock In \emph{Advances in Neural Information Processing Systems},
  volume~30, pages 5998--6008. Curran Associates, Inc.

\bibitem[{Vildov{\'{a}} et~al.(2008)Vildov{\'{a}}, Haji{\v{c}}, Hana,
  Hlav{\'{a}}{\v{c}}ov{\'{a}}, M{\'{i}}rovsk{\'{y}}, and Raab}]{CAC}
Barbora~Hladk{\'{a}} Vildov{\'{a}}, Jan Haji{\v{c}}, Ji{\v{r}}{\'{i}} Hana,
  Jaroslava Hlav{\'{a}}{\v{c}}ov{\'{a}}, Ji{\v{r}}{\'{i}} M{\'{i}}rovsk{\'{y}},
  and Jan Raab. 2008.
\newblock Czech academic corpus 2.0.

\bibitem[{Virtanen et~al.(2019)Virtanen, Kanerva, Ilo, Luoma, Luotolahti,
  Salakoski, Ginter, and Pyysalo}]{finish-bert-2019}
Antti Virtanen, Jenna Kanerva, Rami Ilo, Jouni Luoma, Juhani Luotolahti, Tapio
  Salakoski, Filip Ginter, and Sampo Pyysalo. 2019.
\newblock \href {http://arxiv.org/abs/1912.07076} {Multilingual is not enough:
  Bert for finnish}.

\bibitem[{Vries et~al.(2019)Vries, Cranenburgh, Bisazza, Caselli, Noord, and
  Nissim}]{dutch-BERT}
Wietse~de Vries, Andreas~van Cranenburgh, Arianna Bisazza, Tommaso Caselli,
  Gertjan~van Noord, and Malvina Nissim. 2019.
\newblock \href {http://arxiv.org/abs/1912.09582} {{BERTje}: {A} {Dutch} {BERT}
  {Model}}.
\newblock \emph{arXiv:1912.09582 [cs]}.

\bibitem[{Wu et~al.(2016)Wu, Schuster, Chen, Le, Norouzi, Macherey, Krikun,
  Cao, Gao, Macherey et~al.}]{wu2016google}
Yonghui Wu, Mike Schuster, Zhifeng Chen, Quoc~V Le, Mohammad Norouzi, Wolfgang
  Macherey, Maxim Krikun, Yuan Cao, Qin Gao, Klaus Macherey, et~al. 2016.
\newblock Google's neural machine translation system: Bridging the gap between
  human and machine translation.
\newblock \emph{arXiv preprint arXiv:1609.08144}.

\bibitem[{Yang et~al.(2019)Yang, Dai, Yang, Carbonell, Salakhutdinov, and
  Le}]{nips-2019-XLNET}
Zhilin Yang, Zihang Dai, Yiming Yang, Jaime Carbonell, Russ~R Salakhutdinov,
  and Quoc~V Le. 2019.
\newblock \href
  {https://proceedings.neurips.cc/paper/2019/file/dc6a7e655d7e5840e66733e9ee67cc69-Paper.pdf}
  {Xlnet: Generalized autoregressive pretraining for language understanding}.
\newblock In \emph{Advances in Neural Information Processing Systems},
  volume~32, pages 5753--5763. Curran Associates, Inc.

\bibitem[{Zhao et~al.(2009)Zhao, Chen, Kit, and Zhou}]{zhao2009multilingual}
Hai Zhao, Wenliang Chen, Chunyu Kit, and Guodong Zhou. 2009.
\newblock Multilingual dependency learning: A huge feature engineering method
  to semantic dependency parsing.
\newblock In \emph{Proceedings of the Thirteenth Conference on Computational
  Natural Language Learning (CoNLL 2009): Shared Task}, pages 55--60.

\end{thebibliography}
\bibliographystyle{acl_natbib}

\clearpage

\appendix
\section{Cluster Configuration}
\label{sec:cluster}

We use distributed training to set the weights of Czert. For distributed pre-training we rely on the Czech national cluster Metacentrum\footnote{See \url{https://wiki.metacentrum.cz/wiki/Usage_rules/Acknowledgement}}. We employ 16 machines, each with two NVIDIA TESLA T4 graphic cards, which results in 32 T4s in total.  

For the Czert-A model, we use standard Tensorflow \cite{tensorflow2015-whitepaper} distributed training, which is based upon the gRPC standard. It takes 12 days to training Czert-A with this setting.

The Czert-B model contains almost ten times as many trainable parameters as the Czert-A model. It proved impractical to train Czert-A  with the tools provided by Tensorflow alone. We employ the MPI messaging standard that communicates over the OmniPath network with a speed of 100Gb/s. The Horovod \cite{sergeev2018horovod} library handles all the synchronization transfers of our distributed training. We are able to reach the speeds of 2400ms per batch with this setting, which is approximately five times faster than with standard gRPC via TCP/IP. We are able to train the Czert-B model in 8 days.

\section{Fine-tuning and Hyper-parameters}

\subsection{Named Entity Recognition} \label{token_classification_models}

In all of our experiments, we use Adam optimizer with a learning rate of 5e-5 and a linear decay to zero. Additionally, the Czert-B model uses a learning rate warm-up during the first epoch. All the models are trained with batch size 64 for 25 epochs on an NVIDIA Tesla-T4 GPU. For Czert-A it takes approximately 25 minutes on the \textit{CNEC} dataset, whereas on the \textit{BSNLP 2019} it takes less than 7 minutes.

\subsection{Morphological Tagging}
\label{app:morphtag-hyper}
The architecture of the model follows the token classification setting described in Section \ref{sec:tasks}. The number of output neurons is set to the number of morphological tags in Universal Dependencies. Namely:
\begin{itemize}
\setlength\itemsep{\itemizeseparatorsize}
\item Prague Dependency Treebank 3.0,
\item Czech Academic Corpus 2.0,
\item Czech Legal Text Treebank 2.0,
\item FicTree.
\end{itemize}

For fine-tuning, we use Adam optimizer with a learning rate of 5e-5 and a linear decay to zero. Additionally, the Czert-B model uses a learning rate warm-up during the first epoch. Similarly to our NER experiments (Section \ref{sec:T_NER}s), we use a maximum sequence length of 128 sub-word tokens. The models are trained with batch size 64 for 13 epochs. For Czert-A it takes about 8 hours and 15 minutes on an NVIDIA Tesla-T4 GPU.

\subsection{Semantic Role Labeling} \label{app:srl}

For fine-tuning, we use Adam optimizer with a learning rate of 5e-5 and a linear decay to zero. We use a maximum sequence length of 128 sub-word tokens. We train the model on 2 Tesla T4 graphic cards with batch size of 64 for 12 epochs.

\subsection{Sentiment Classification} \label{app:sentiment}

\par We perform fine-tune training of the models by minimizing the Cross-Entropy loss function using the Adam \cite{Kingma-adam} optimization algorithm with default parameters ($\beta_1 = 0.9, \beta_2 = 0.999$) and with a linear learning rate decay (without warm-up). We try three different initial learning rates, i.e.,  2e-6,  2e-5 and  2.5e-5 for at most 14 epochs. We use a max sequence length of 64, batch size of 32 for the FB\footnote{Even though that we use different tokenizers for each model, number of tokens in posts from the FB dataset do not exceed 66 tokens and average number of tokens around 20 for all tokenizers.} dataset and a max sequence length of 512 and batch size of 14 for the CSFD dataset.

\subsection{Semantic Textual Similarity}
\label{app:sts}
\label{sec:appendix}

For the CNA dataset, we train two epochs using a batch of size 50, and LR 1e-5 with linear decay to zero for each model except Czert-A for which we used 5e-6 for four epochs, which lead to slightly better results. 

For smaller datasets (SVOB-img and SVOB-hl) we used LR 5e-6 and train on 14k batches.

For each experiment, we used Adam optimizer, L2 weight normalization, and learning rate warm-up during the first 500 batches.

\subsection{Multi-label Document Classification} \label{app:mlc}
For each experiment, we first run a linear grid search through learning rate parameter L = \{2e-5, 4e-5, ..., 10e-4\} and a decision $D = \{true, false\}$  whether to use a linear learning rate decay\footnote{Arriving at 0 at the end of the last epoch.} or to keep the maximum learning rate constant until the last step. The learning rate achieved maximum after 500 steps of the warm-up phase. After the grid search was complete, we've run twenty 10-epoch-long training phases for each of the extended models and average the results. 

\end{document}